\documentclass[runningheads]{llncs}
\usepackage{graphicx}
\usepackage{comment}
\usepackage{amsmath,amssymb} 
\usepackage{color}
\usepackage{footnote}

\usepackage[colorlinks]{hyperref}
\usepackage[sort, numbers]{natbib}

\usepackage{enumitem}
\usepackage{tablefootnote}
\usepackage{multirow}
\usepackage{capt-of}
\usepackage{fixltx2e}
\usepackage{floatrow}
\usepackage{pifont}

\def\etal{\emph{et al.}}
\def\etc{\emph{etc}}
\def\ie{\emph{i.e}}

\begin{document}
\pagestyle{headings}
\mainmatter
\def\ECCVSubNumber{1308}  

\title{Learning to Count in the Crowd from Limited Labeled Data} 

%
%

\author{Vishwanath A. Sindagi\inst{1} \and
Rajeev Yasarla\inst{1}\and
Deepak Sam Babu\inst{2}\and
R. Venkatesh Babu\inst{2}\and
Vishal M. Patel\inst{1}
}

\authorrunning{Sindagi et al.}
%
\institute{Johns Hopkins University, Baltimore MD 21218, USA \and
Indian Institute of Science, Bangalore 560012, India\\
\email{\{vishwanathsindagi,ryasarl1,vpatel36\}@jhu.edu}
\email{\{deepaksam,venky\}@iisc.ac.in}}
\maketitle

\begin{abstract}
Recent crowd counting approaches have  achieved excellent performance.  However, they are essentially based on fully supervised paradigm and   require  large number of annotated  samples. Obtaining annotations is an  expensive and labour-intensive process. In this work, we focus on reducing the annotation efforts by learning to count in the crowd from limited number of labeled samples while leveraging a large pool of unlabeled data. Specifically, we propose a Gaussian Process-based iterative learning mechanism that involves  estimation of pseudo-ground truth for the unlabeled data, which is then used as supervision for training the network. The proposed method is shown to be effective   under the reduced data  (semi-supervised) settings for several  datasets like ShanghaiTech, UCF-QNRF, WorldExpo, UCSD, \etc. Furthermore, we demonstrate that the proposed method can be  leveraged to enable the network in learning to count from synthetic dataset while being able to generalize better to real-world datasets (synthetic-to-real transfer). 
 

\end{abstract}

\section{Introduction}
Due to its significance in several applications (like video surveillance \cite{kang2017beyond,toropov2015traffic,sindagi2019dafe}, public safety monitoring \cite{zhan2008crowd}, microscopic cell counting \cite{lempitsky2010learning}, environmental studies \cite{lu2017tasselnet}, \etc.), crowd counting has attracted a lot of interest from the deep learning research community.  Several convolutional neural network (CNN) based approaches have been developed that address various issues in counting like scale variations, occlusion, background clutter \cite{li2015crowded,zhang2015cross,li2014anomaly,sam2017switching,sindagi2017generating,liu2018leveraging,shi2018crowd_negative,shen2018adversarial,cao2018scale,ranjan2018iterative,li2018csrnet,sindagi2017survey,sam2019locate,sam2020going,babu2018divide}, \etc.  While these methods have achieved excellent improvements  in terms of the overall error rate,  they follow a fully-supervised paradigm and  require several labeled data samples. There is a wide variety of scenes and crowded scenarios that these networks need  to handle to in the real world.  Due to a distribution gap between the training and testing environments, these networks have limited generalization abilities and hence,  procuring annotations becomes especially important. However, annotating data for crowd counting typically involves obtaining point-wise annotations at head locations, and this is a labour intensive and expensive process.  Hence,  it is infeasible to procure annotations for all possible scenarios. Considering this, it is crucial to reduce the annotation efforts, especially for   crowd counting methods which get deployed in a wide variety of scenarios. 

\begin{figure}[t!]
	\begin{center}
		\includegraphics[width=.91\linewidth]{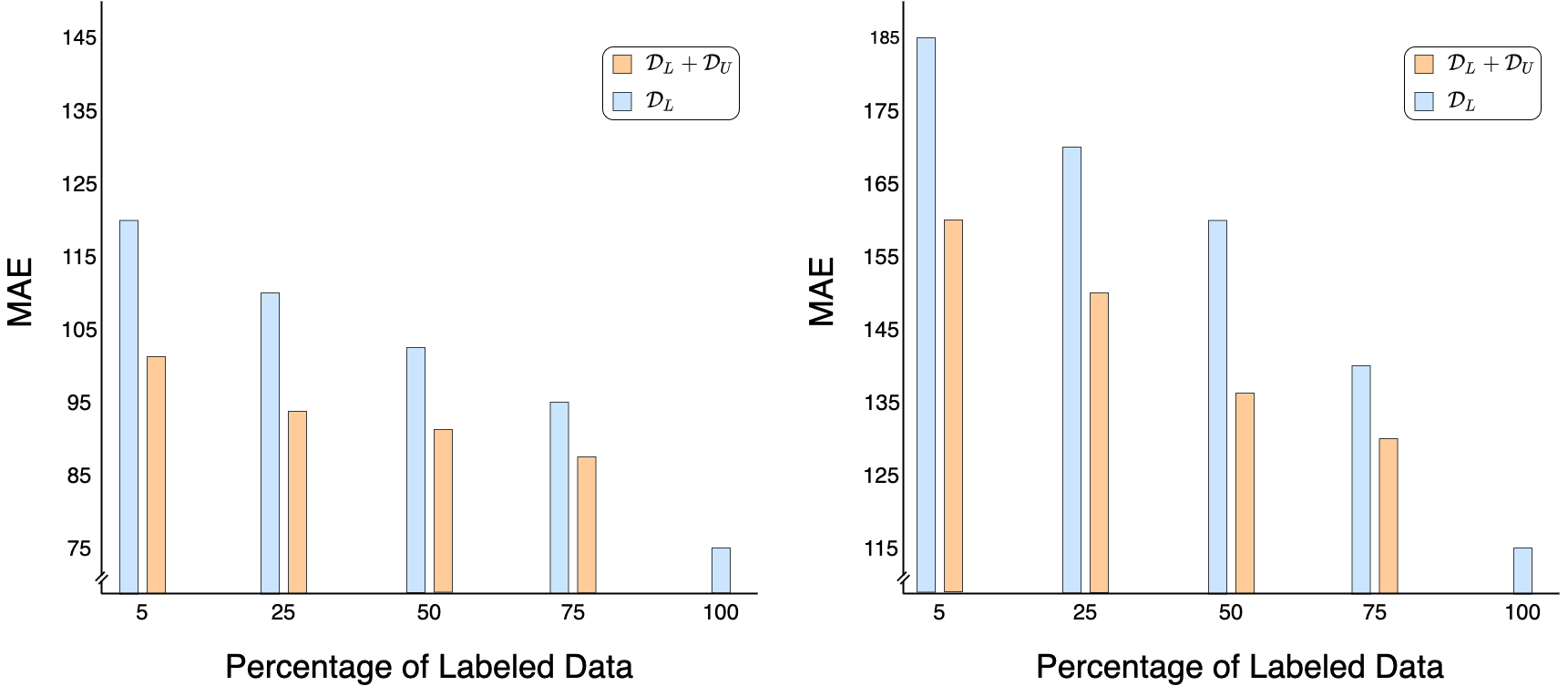} 
	\end{center}
	\vskip -8pt
	(a) \hskip 100pt (b)
	\vskip -8pt \caption{Results of semi-supervised learning experiments. (a) ShanghaiTech A (b) UCF-QNRF. For both datasets, the error increases with reduction in the \%-age of labeled data. By leveraging the unlabeled dataset using the proposed GP-based framework, we are able to reduce the error considerably. Note that $\mathcal{D_L}$ and $\mathcal{D_U}$ indicate labeled and unlabeled dataset, respectively.}
	\label{fig:graph}
\end{figure}

With the exception of a few works \cite{change2013semi,liu2018leveraging, wang2019learning}, reducing annotation efforts while maintaining good performance is relatively less explored for the task of crowd counting. Hence, in this work, we focus on  learning to count using limited labeled data  while leveraging unlabeled data to improve the performance. Specifically, we propose a Gaussian Process (GP) based iterative  learning framework where we augment the existing networks with capabilities to leverage unlabeled data, thereby resulting in overall improvement in the performance. The proposed framework follows a pseudo-labeling approach, where we estimate the pseudo-ground truth (pseudo-GT) for the unlabeled data, which is then used to supervise the network. The network is trained iteratively on labeled and unlabeled data. In the labeled stage, the network weights are updated by minimizing the $L_2$ error between predictions and the ground-truth (GT) for the labeled data. In addition, we save the latent space vectors of the labeled data along with the ground-truths.  In the  unlabeled  stage, we first  model the relationship between  the latent space vectors of the labeled  images along with the corresponding ground-truth    and unlabeled latent space vectors  jointly using GP. Next, we estimate the  pseudo-GT for the unlabeled inputs using the GP modeled earlier. This pseudo-GT is then used to supervise the network for the unlabeled data.  Minimizing the error between the unlabeled data predictions and the pseudo-GT results in improved performance. Fig. \ref{fig:graph} illustrates the effectiveness of the proposed GP-based framework in exploiting unlabeled data on two datasets (ShanghaiTech-A \cite{zhang2016single} and UCF-QNRF\cite{idrees2018composition}) in the reduced data setting. It can be observed that the proposed method is able to leverage unlabeled data effectively resulting in lower error across various settings. 

The proposed method is evaluated on different datasets like ShanghaiTech \cite{zhang2016single}, UCF-QNRF \cite{idrees2018composition}, WorldExpo \cite{zhang2015cross}, UCSD \cite{chan2008privacy}, \etc. in the reduced data settings. In addition to obtaining lower error as compared to the existing methods \cite{liu2018leveraging},  the performance drop due to less data is improved by a considerable margin. Furthermore, the proposed method is effective for learning to count from synthetic data as well. More specifically, we use labeled synthetic crowd counting dataset (GCC \cite{wang2019learning}) and unlabeled real-world datasets (ShanghaiTech \cite{zhang2016single}, UCF-QNRF \cite{idrees2018composition}, WorldExpo \cite{zhang2015cross}, UCSD \cite{chan2009bayesian}) in our framework, and show that it is able to generalize better to real-world datasets as compared to recent domain adaptive crowd counting approaches \cite{wang2019learning}. To summarize, the following are our contributions:
 \begin{itemize}[topsep=0pt,noitemsep,leftmargin=*]
	\item We propose a GP-based framework to effectively exploit unlabeled data during the training process, resulting in improved  overall performance. The proposed method consists of iteratively training over labeled and unlabeled data. For the unlabeled data, we estimate the pseudo-GT using the GP modeled during labeled phase. 
	\item We demonstrate that the proposed framework is effective in semi-supervised and synthetic-to-real transfer settings. Through various ablation studies, we show that the proposed method is generalizable to different network architectures and various reduced data settings.  
\end{itemize}

\section{Related Work}

\noindent\textbf{Crowd Counting.} Traditional approaches in crowd counting  (\cite{li2008estimating,ryan2009crowd,chen2012feature,idrees2013multi,lempitsky2010learning,pham2015count,xu2016crowd}) typically involved feature extraction techniques and training regression algorithms. Recently, CNN-based approaches like \cite{wang2015deep,zhang2015cross,sam2017switching,arteta2016counting,walach2016learning,onoro2016towards,zhang2016single,sam2017switching,sindagi2017generating} have surpassed the traditional approaches by a large margin in terms of the overall error rate. Most of these methods focus on addressing the issue of  large variations in scales. Approaches like \cite{zhang2016single,sam2017switching,sindagi2017generating} focus on improving the receptive field.  Different from these,  approaches like \cite{ranjan2018iterative,sindagi2017cnnbased,sindagi2019multi,sam2018top} focus on effective ways of fusing multi-scale information from deep networks. In addition to scale variation, recent approaches have addressed other issues in crowd counting like improving the quality of predicted density maps using adversarial regularization \cite{sindagi2017generating, shen2018adversarial}, use of deep negative correlation-based learning for obtaining more generalizable  features, and scale-based feature aggregation \cite{cao2018scale}. Most recently, several methods have employed additional information like   segmentation and semantic priors
\cite{zhao2019leveraging,wan2019residual}, attention \cite{liu2018adcrowdnet,sindagi2019ha,sindagi2019inverse}, perspective \cite{shi2019revisiting}, context information  \cite{liu2019context}, multiple-views \cite{zhang2019wide}  and multi-scale features  \cite{jiang2019crowd}, adaptive density maps \cite{wan2019adaptive} into the network. 
In other efforts,  researchers have made important contributions by creating large-scale datasets for  counting like UCF-QNRF \cite{idrees2018composition}, GCC  \cite{wang2019learning}  and JHU-CROWD \cite{sindagi2019pushing,sindagi2020jhu}. For a more detailed discussion on these methods, the reader is referred to recent comprehensive surveys \cite{sindagi2017survey,gao2020cnn}



\noindent\textbf{Learning from limited data. } Recent research in crowd counting has been largely focused on improving the counting performance in the fully-supervised paradigm. Very few works  like \cite{change2013semi,  liu2018leveraging,wang2019learning} have made efforts on minimizing annotation efforts. Loy \etal \cite{change2013semi} proposed a semi-supervised regression framework  that exploit underlying geometric structures of crowd patterns to assimilate the count estimation of two nearby crowd pattern points in the manifold. However, this approach is specifically designed for video-based crowd counting. 

Recently, Liu \etal \cite{liu2018leveraging} proposed to leverage additional unlabeled data for counting by introducing a learning to rank framework. They assume that  any sub-image of a crowded scene image is guaranteed to contain the same number or fewer persons than the super-image. They employ pairwise ranking hinge loss to enforce this ranking constraint for unlabeled data in addition to the $L_2$ error to train the network. In our experiments we observed that this constraint is almost always satisfied, and it provides relatively less supervision over unlabeled data.

Babu \etal \cite{sam2019almost} focus on a different approach, where they train 99.9\% of their parameters from unlabeled data using a novel unsupervised learning framework based on winner-takes-all (WTA) strategy.  However, they still train the remaining set of parameters using labeled data.


Wang \etal \cite{wang2019learning} take a totally different approach to minimize annotation efforts by creating a new synthetic crowd counting dataset (GCC).   Additionally, they propose a Cycle-GAN based domain adaptive approach for generalizing the network trained on synthetic dataset to real-world dataset. However,there is a large gap  in terms of the style and also the crowd count between the synthetic and real-world scenarios.  Domain adaptive approaches have limited abilities in handling such scenarios. In order to obtain successful adaptation,  the authors in \cite{wang2019learning} manually select the samples from the synthetic dataset that are closer to the real-world scenario in terms of crowd count for training the network. This selection is possible when one has information about the count from the real-world datasets, which violates the assumption of lack of unlabeled data in the target domain for unsupervised domain adaptation. 

Considering the drawbacks of existing approaches, we propose a new GP-based iterative training framework to exploit unlabeled data. 

\section{Preliminaries}

In this section, we briefly review  the concepts (crowd counting, semi-supervised learning and Gaussian Process) that are used in this work. \\

\noindent\textbf{Crowd counting.}
Following recent works \cite{zhang2015cross,zhang2016single}, we employ the approach of  density estimation technique. That is, an input crowd image is forwarded through the network, and the network outputs a density map. This density map indicates the per-pixel count of people in the image. The count in the image is obtained by integrating over the density map.  For training the network using labeled data, the ground-truth density maps are  obtained by imposing 2D Gaussians at head location $x_g$  using $D(x) = \sum_{{x_g \in S}}\mathcal{N}(x-x_g,\sigma)$. Here,   $\sigma$ is the Gaussian kernel's scale and $S$ is the list of all locations of people.   \\
 
\noindent\textbf{Problem formulation.}
We are given a set of labeled dataset of input-GT pairs ($\{x,y\} \in \mathcal{D_L}$)  and a set of unlabeled input data samples $x \in \mathcal{D_U}$. The objective is to fit a  mapping-function $f(x|\phi)$  
(with parameters defined by $\phi$) that accurately estimates  target label $y$ for unobserved samples.  Note that this definition applies to both semi-supervised setting and synthetic-to-real transfer  setting. In the case of synthetic-to-real transfer, the synthetic dataset is labeled and hence, can be used as the labeled dataset ($\mathcal{D_L}$). Similarly, the real-world  dataset is unlabeled and can be used as the unlabeled dataset ($\mathcal{D_U}$).

In order to learn the parameters, both labeled and unlabeled datasets are exploited. Typically, 
loss functions such as $L_1$,  $L_2$ or cross entropy  error are used for labeled data. For exploiting unlabeled data $\mathcal{D_U}$, existing approaches augment $f(x|\phi)$ with information like shape of the data manifold \cite{oliver2018realistic} via different techniques such as enforcing consistent regularization \cite{laine2016temporal}, virtual adversarial training \cite{miyato2018virtual}  or pseudo-labeling \cite{lee2013pseudo}. In this work, we employ pseudo-labeling based approach where we estimate pseudo-GT for unlabeled data, and then use them for supervising the network using traditional supervised loss functions. \\

\setlength{\belowdisplayskip}{0pt} \setlength{\belowdisplayshortskip}{0pt}
\setlength{\abovedisplayskip}{0pt} \setlength{\abovedisplayshortskip}{0pt}

\noindent\textbf{Gaussian process.}
A Gaussian process (GP) $f(v)$ is an infinite collection of random variables, any finite subset of which have a joint Gaussian distribution. A GP is fully specified by its mean function ($m(v)$) and covariance function $K(v,v')$. These are defined below:   
\begin{equation}
\begin{aligned} m(v) &=\mathbb{E}[f(v)],
\end{aligned}
\end{equation}
\begin{equation}
\begin{aligned} {K}\left(v, v^{\prime}\right) &=\mathbb{E}\left[(f(v)-m(v))\left(f\left(v^{\prime}\right)-m\left(v^{\prime}\right)\right)\right], \end{aligned}
\end{equation}
where $v,v' \in \mathcal{V}$ denote the possible inputs that index the GP. The covariance matrix is computed  from the covariance function ${K}$ which expresses the notion of smoothness of the underlying function. GP can then be formulated as follows:
\begin{equation}
f(v) \sim \mathcal{GP}(m(v), K(v, v')+\sigma_{\epsilon}^2I), 
\end{equation}
where  $I$ is identity matrix and $\sigma_{\epsilon}^2$ is the variance of the additive noise.  Any collection of function values is then jointly Gaussian as follows
\begin{equation}
f(V)=\left[f\left(v_{1}\right), \ldots, f\left(v_{n}\right)\right]^{T} \sim \mathcal{N}\left(\mu, K(V, V')+\sigma_{\epsilon}^2I\right),
\end{equation}
with mean vector and covariance matrix defined by the GP as mentioned earlier. To make predictions at unlabeled points, one can compute a Gaussian posterior distribution in closed form by conditioning on the observed data. For  more details, we refer the reader  to \cite{rasmussen2003gaussian}.

\section{GP-based iterative learning}
\begin{figure}[t!]
	\begin{center}
		\includegraphics[width=0.95\linewidth]{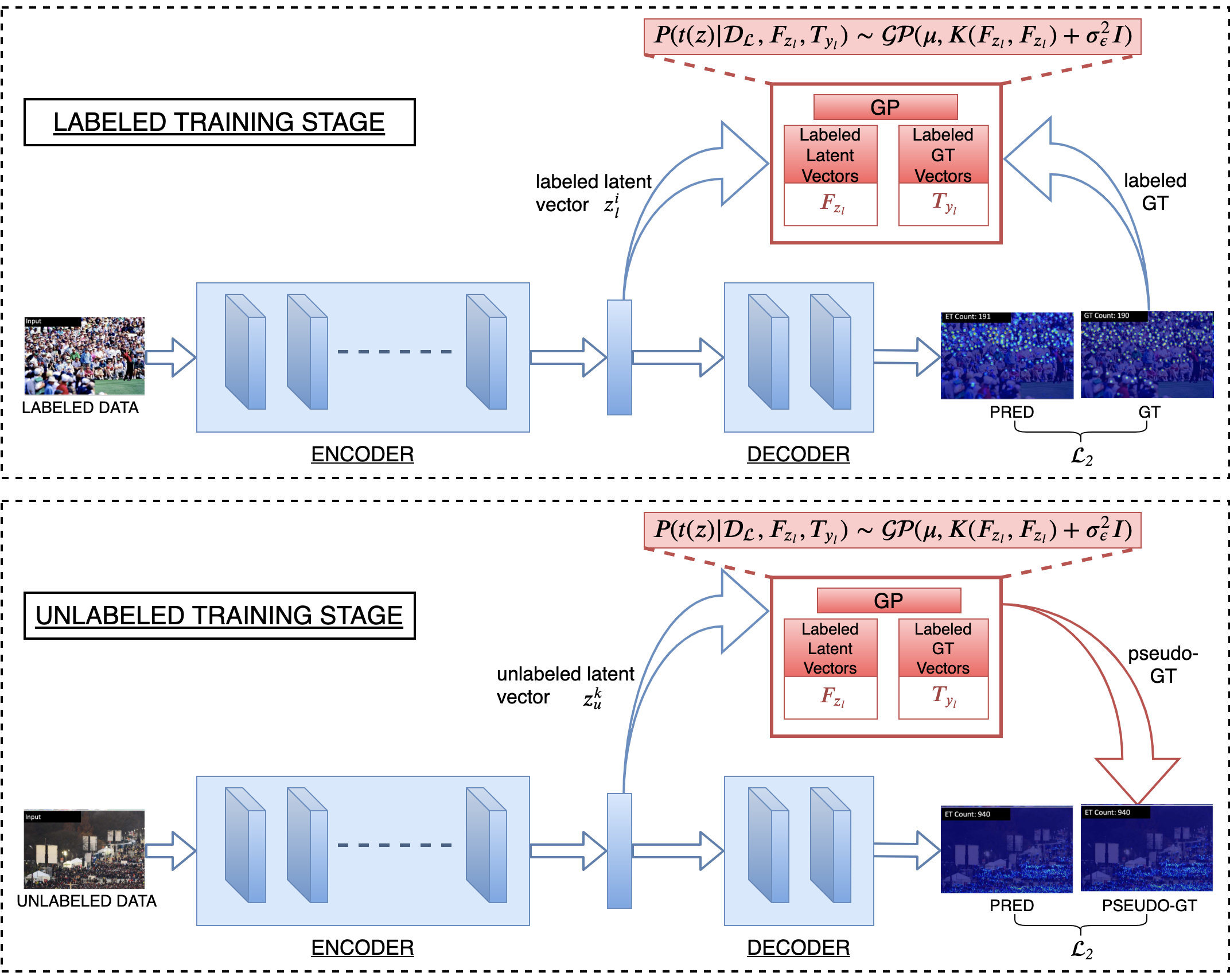} 
	\end{center}
	\vskip -15pt \caption{Illustration of the proposed framework. Training is performed iteratively over labeled and unlabeled data. For labeled data, we minimize the $L_2$ error between the predictions and GT.  For unlabeled data, we  minimize the $L_2$ error between the predictions and pseudo-GT.}
	\label{fig:overview}
\end{figure}

Fig. \ref{fig:overview} gives an overview of the proposed method. The network is constructed using  an encoder  $f_{e}(x,\phi_{e})$  and  a decoder $f_{d}(z,\phi_{d})$, that are parameterized by $\phi_{e}$ and $\phi_{d}$, respectively. The proposed framework is agnostic to the encoder network, and we show in the experiments section that it generalizes well to architectures such as VGG16 \cite{simonyan2014very}, ResNet-50 and ResNet-101 \cite{ren2015faster}. The decoder consists of a set of 2 conv-relu layers (see supplementary material for more details). Typically, an input crowd image $x$ is forwarded through the encoder network to obtain the corresponding latent space vector $z$.  This vector is then forwarded through the decoder network to obtain  the crowd density output $y$, \ie, $y = f_d(f_e(x,\phi_{e}),\phi_d)$.

We are given a training dataset, $\mathcal{D}=\mathcal{D_L} \cup \mathcal{D_U}$, where $\mathcal{D_L}=\{x_l^i,y_l^i\}_{i=1}^{N_l}$ is a labeled dataset containing  $N_{l}$ training samples and $\mathcal{D_U}=\{x_u^i\}_{i=1}^{N_u}$ is an  unlabeled dataset containing  $N_{u}$ training samples.  The proposed framework effectively leverages both the datasets by iterating the training process over labeled $\mathcal{D_L}$ and unlabeled datasets $\mathcal{D_U}$. More specifically, the training process consists of two stages: (i) Labeled training stage: In this stage, we employ supervised loss function $\mathcal{L}_{s}$  to learn the network parameters using labeled dataset, and (ii) Unlabeled training stage: We generate pseudo GTs for the unlabeled data points using the GP formulation, which is then used for supervising the network on the unlabeled dataset.  In what follows, we describe these stages in detail. 

\subsection{Labeled  stage}
Since the labeled dataset  $\mathcal{D_L}$ comes with annotations, we employ $L_2$ error between the predictions and the GTs as supervision loss for training the network. This loss objective  is defined as follows:
\begin{equation}
\label{eq:loss_sup}
\mathcal{L}_{s} = \mathcal{L}_2  =  \|y^{pred}_l - y_l\|_2,
\end{equation}
where $y^{pred}_l=g(z_l,\phi_{d})$ is the predicted output,  $y_l$ is the ground-truth,  $z=h(x, \phi_{e})$ is the intermediate latent space vector. Note that, the subscript $l$ in the above quantities indicate that these are defined for labeled data. 

Along with performing supervision on the labeled data, we additionally save   feature vectors  $z_l^i$'s  from the intermediate latent space in  a matrix $F_{z_l}$. Specifically, $F_{z_l}= \{z_l^i\}_{i=1}^{N_l}$. This matrix is used for computing the pseudo-GTs for unlabeled data at a later stage. The dimension of  $F_{z_l}$  matrix is  $N_l\times M$. Here, $M$ is the dimension of the latent space vector $z_l$. In our case, the latent space vector dimension is $64\times32\times32$ (see supplementary material for more details), which is reshaped to $1\times65,536$. Hence,  $M=65,536$.

\subsection{Unlabeled  stage}
Since the unlabeled data $\mathcal{D_U}$ does not come with any GT annotations, we estimate pseudo-GTs which are then used as supervision for training the network on unlabeled data. For this purpose, we  model the relationship between  the latent space vectors of the labeled  images $F_{z_l}$ along with the corresponding GT $T_{y_l}$ and unlabeled latent space vectors $z^{pred}_u$  jointly using GP.\\

\noindent\textbf{Estimation of pseudo-GT:} As discussed earlier, the training process iterates over labeled  $\mathcal{D_L}$ and unlabeled data $\mathcal{D_U}$.  After the labeled stage, the labeled latent space vectors $F_{z_l}$  and their corresponding GT density maps $T_{y_l}$   are used to  model the function $t$ which maps the relationship between the  latent vectors and  the  output density maps as, $y = t(z)$.  Using GP, we  model this function $t(.)$ as an infinite collection of functions of which any finite subset is jointly Gaussian.  More specifically, we  jointly model the distribution of the function values $t(.)$ of the latent space vectors of the labeled and the unlabeled samples using GP as follows:
\begin{equation}
P(t(z)|\mathcal{D_L},F_{z_l},T_{y_l}) \sim \mathcal{GP}(\mu, K(F_{z_l},F_{z_l})+\sigma_{\epsilon}^2I),
\end{equation}
where $\mu$ is the function value computed using GP, $\sigma_\epsilon^2$ is set equal to 1, and $K$  is the kernel function. Based on this, the conditional joint distribution  for the latent space vector   $z^k_u$ of the  $k^{th}$ unlabeled sample $x^k_u$ can be expressed as the following Gaussian distribution:
%
\begin{equation}
P(t(z_u^k)|\mathcal{D_L},F_{z_l},T_{z_l}) = \mathcal{N}(\mu_u^k,\Sigma_u^k),
\label{eq:cond_prob}
\end{equation}
where 
\begin{equation}
\label{eq:mean}
\mu_u^k = K(z_u^k, F_{z_l}) [K(F_{z_l},F_{z_l}) + \sigma_\epsilon^2 I]^{-1}T_{y_l},
\end{equation}
\vskip2pt
\begin{equation}
\label{eq:sigma}
\begin{aligned}
\Sigma_u^k = {} & K(z_u^k,z_u^k) - K(z_u^k,F_{z_l})[K(F_{z_l},F_{z_l})+\sigma_\epsilon^2I]^{-1}  K(F_{z_l},z_u^k) + \sigma_\epsilon^2
\end{aligned}
\end{equation}
where $\sigma_\epsilon^2$ is set equal to 1 and $K$  is a kernel function with the following definition:
\begin{equation}
K(Z,Z)_{k,i}= \kappa(z_u^k,z_l^i) = \frac{ \langle z_u^k, z_l^i\rangle}{|z_u^k|\cdot|z_l^i|}.
\end{equation}

Considering the large dimensionality of the latent space vector,  $K(F_{z_l},F_{z_l})$ can grow quickly in size especially if the number of labeled data samples $N_l$ is high. In such cases, the computational and memory requirements become prohibitively high. Additionally, all the latent vectors may not be  necessarily effective since these vectors correspond to different regions of images in terms of content and size/density of the crowd. In order to overcome these issues, we use only those labeled vectors that are similar to the unlabeled latent vector. Specifically, we consider only $N_n$ nearest labeled vectors corresponding to an unlabeled vector. That is, we replace  $F_{z_l}$ by $F_{z_l,n}$ in Eq. \eqref{eq:cond_prob}-\eqref{eq:sigma}. Here $F_{z_l,n}=\{z_l^j :  z_l^j \in nearest(z_u^k,F_{z_l} ,N_n) \}$, and $T_{y_l,n}=\{y^j_l:  z_l^j \in nearest(z_u^k,F_{z_l} ,N_n) \}$ with $nearest(p,Q ,N_n)$ being a function that finds top $N_n$ nearest neighbors of $p$ in $Q$.  

The pseudo-GT for unlabeled data sample is given by the mean predicted in  Eq. \eqref{eq:mean}, \ie, $y_{u,pseudo}^{k} = \mu_u^k$. The $L_2$ distance between the predictions $y^{k}_{u,pred}=g(z_u^k,\phi_{e})$ and the pseudo-GT $y_{u,pseudo}^{k}$ is used as supervision for updating the parameters of the encoder $f_e(\cdot,\phi_{e})$ and the decoder $f_d(.,\phi_{d})$.

Furthermore, the pseudo-GT estimated using Eq. \eqref{eq:mean} may not be necessarily perfect. Errors in pseudo-GT will limit the performance of the  network. To overcome this, we explicitly exploit the variance modeled by the GP. Specifically,  we  minimize the predictive  variance by considering   Eq. \eqref{eq:sigma} in  the loss function. As discussed earlier, using all the latent space vectors of labeled data may not be necessarily effective. Hence, we minimize the variance $\Sigma_{u,n}^k$ computed between $z^{k}_{u}$ and the $N_n$ nearest neighbors in the latent space vectors using GP.  Thus, the loss function during the unlabeled stage is defined as: 
\begin{equation}
\label{eq:loss_unsup}
\mathcal{L}_{un} = \frac{1}{|\Sigma_{u,n}^{k}|}\|{y}^{k}_{u,pred} - {y}_{u,pseudo}^{k}\|_2 + \log \Sigma_{u,n}^{k},
\end{equation}
where $y^{k}_{u,pred}$ is the crowd density map prediction obtained by forwarding an unlabeled input image $x_u^k$ through the network, $y_{u,pseudo}^{k} = \mu_u^k$ is the  pseudo-GT (see Eq. \eqref{eq:mean}), and $\Sigma_{u,n}^{k}$ is the predictive variance obtained by replacing $F_{z_l}$ in Eq. \eqref{eq:sigma} with $F_{z_l,n}$.

\subsection{Final objective function}
We combine the supervised loss  Eq. \eqref{eq:loss_sup} and unsupervised loss Eq. \eqref{eq:loss_unsup} to obtain the final objective function as follows:
\begin{equation}
\label{eq:loss_total}
\mathcal{L}_{f} = \mathcal{L}_{s} + \lambda_{un} \mathcal{L}_{un},
\end{equation} 
where $\lambda_{un}$ is a hyper-parameter that weighs the unsupervised loss.

\section{Experiments and results}
In this section, we discuss the details of the various experiments conducted to demonstrate the effectiveness of the proposed method.  Since the proposed method is able to leverage unlabeled data to improve the overall performance, we performed evaluation in two settings: (i) \textit{Semi-supervised settings}: In this setting, we varied  the percentage of labeled samples from 5\% to 75\%. We first show that with the base network, there is performance drop due to the reduced data. Later, we show that the proposed method is able to recover a major percentage of the performance drop.  (ii) \textit{Synthetic-to-real transfer  settings}: In this setting, the goal is to train on synthetic dataset (labeled), while adapting to real-world dataset. Unlabeled images from the real-world are available during  training. In both settings, the proposed method is able to achieve better results as compared to recent methods.   Details of the datasets are provided in the supplementary material.

\subsection{Semi-supervised settings}
In this section, we conduct experiments in the semi-supervised settings by reducing the amount of labeled data available during training. The rest of the samples in the dataset are considered as unlabeled samples wherever applicable. In the following sub-sections, we present comparison of the proposed method in the 5\% setting with other recent methods.  For comparison, we used 4 datasets: ShanghaiTech (SH-A/B)\cite{zhang2016single}, UCF-QNRF \cite{idrees2018composition}, WorldExpo \cite{zhang2015cross} and UCSD \cite{chan2008privacy}. This is followed by  a detailed ablation study involving different architectures and various percentages of labeled data used during training. For ablation, we chose ShanghaiTech-A and UCF-QNRF datasets since they contain a wide diversity of scenes and large variation in count and scales. \\

\noindent\textbf{Implementation details.} We train the network using Adam optimizer with a learning rate of $10e-{5}$  and a momentum of $0.9$ on an NVIDIA Titan Xp GPU. We use batch size of 24. During training, random crops of size $256\times256$ are used. During inference, the entire image is forwarded through the network. For evaluation, we use mean absolute error ($MAE$) and mean squared error ($MSE$) metrics, which are defined as: $MAE = \frac{1}{N}\sum_{i=1}^{N}|y_i-y'_i|$ and  $MSE = \sqrt{\frac{1}{N}\sum_{i=1}^{N}|y_i-y'_i|^2}$, respectively.  Here,  $N$ is the total number of test images, $y_i$ is the ground-truth/target count of people in the image and $y'_i$ is the predicted count of people  in to the $i^{th}$ image. We set aside 10\% of the training set for the purpose of validation. The hyper-parameter $\lambda_{un}$ was chosen based on the validation performance. More details are provided in the supplementary.\\

\begin{table}[t!]
	
	\centering
	\caption{Comparison of results in SSL settings.  Reducing labeled data to 5\% results in performance drop by a big margin as compared to 100\% data. ResNet-50 was used as the encoder network for all the methods. RL: Ranking-Loss. GP: Gaussian-Process. AG: Average Gain \%$^{\ref{ftn:ag}}$.}
	\label{tab:ssl_compare}
	\resizebox{1\linewidth}{!}{
	\begin{tabular}{l|c|c|ccc|ccc|ccc|cc|ccc} 
		\hline
		\multirow{2}{*}{Method} & \multirow{2}{*}{$\mathcal{D_L}$ } & \multirow{2}{*}{$\mathcal{D_U}$ } & \multicolumn{3}{c|}{SH-A}            & \multicolumn{3}{c|}{SH-B}              & \multicolumn{3}{c|}{UCF-QNRF}        & \multicolumn{2}{c|}{WExpo} & \multicolumn{3}{c}{UCSD}             \\ 
		\cline{4-17}
		&                                   &                                   & MAE           & MSE           & AG & MAE            & MSE            & AG& MAE           & MSE           & AG & MAE            & AG     & MAE           & MSE           & AG \\ 
		\hline
		ResNet-50 (Oracle)          & 100\%                             & -                                 & 76            & 126           & -    & 8.4            & 14.5           & -    & 114           & 195           & -    & 10.1           & -         & 1.7           & 2.1           & -     \\ 
		\hline\hline
		ResNet-50 ($\mathcal{D_L}$-only)                  & 5\%                               & -                                 & 118           & 211           & -    & 21.2           & 34.2           & -    & 186           & 295           & -    & 14.2           & -         & 2.2           & 2.8           & -     \\
		ResNet-50+RL                & 5\%                               &            95\%                       & 115           & 208           & 2.0   & 20.1           & 32.9           & 4.0    & 182           & 291           &  1.7  & 14.0           &  0.01        & 2.2           & 2.8           &  0     \\
		ResNet-50+GP(Ours)          & 5\%                               &           95\%                         & \textbf{102}  & \textbf{172}  & \textbf{16}    & \textbf{15.7}  & \textbf{27.9}  & \textbf{22}     & \textbf{160}  & \textbf{275}  &  \textbf{10}     & \textbf{12.8}  & \textbf{10}         & \textbf{2.0}  & \textbf{2.4}  &  \textbf{12}     \\
		\hline
	\end{tabular}
}
\end{table}

\noindent\textbf{Comparison with recent approaches.} 
Here, we compare  the effectiveness of the proposed method with a recent method by Liu \etal \cite{liu2018leveraging} on 4 different datasets.  In order to get a better understanding of the overall improvements, we also provide the results of the base network with (i) 100\% labeled data supervision that is the oracle performance, and (ii) 5\% labeled data supervision.

For all the methods (except oracle), we limited the labeled data used during training to 5\% of the training dataset. Rest of the samples were used as unlabeled samples. We used ResNet-50 as the encoder network. The results of the experiments are shown in Table \ref{tab:ssl_compare}. For all the  experiments that we conducted, we report the average of the results for 5 trials. The standard deviations are reported  in the supplementary. We make the following observations for all the datasets: (i) Compared to using the entire dataset, reducing the labeled data during training (to 5\%)  leads to significant increase in error. (ii) The proposed GP-based framework is  able 
to reduce the performance drop by a large margin. Further, the proposed method achieves an average gain (AG)\footnote{\label{ftn:ag}$AG=\frac{G_{mae}+G_{mse}}{2}$, $G_{mae} =  \frac{mae_{(\mathcal{D_U+D_L})} - mae_{(\mathcal{D_L})}}{mae_{(\mathcal{D_L})}}$, $G_{mse} =  \frac{mse_{(\mathcal{D_U+D_L})} - mse_{(\mathcal{D_L})}}{mse_{(\mathcal{D_L})}}$} of anywhere between 10\%-22\% over the $\mathcal{D_L}$-only baseline across all datasets. (iii) The proposed method is able to leverage the unlabeled data more effectively as compared to Liu \etal \cite{liu2018leveraging}. This is because the authors in \cite{liu2018leveraging} using a ranking loss on the unlabeled data which is based on the assumption that  sub-image of a crowded scene is guaranteed to contain the same or fewer number of people compared to the entire image. We observed that this constraint is satisfied naturally  for most of the unlabeled images, and hence it provides less supervision (see supplementary material for a detailed analysis).\\

\begin{table}[t!]
	\caption{Results of ablation study with different \%-ages of labeled data. The proposed method achieves significant gains across different percentages of labeled data. We used ResNet-50 as the encoder network for all the experiments. AG: Average Gain \%$^{\ref{ftn:ag}}$. }
	\label{tab:ablation_lab}
	\vskip-9pt
	\resizebox{.81\linewidth}{!}{
	\begin{tabular}{l|cc|cc|c|cc|cc|c}
		\hline
		\multirow{3}{*}{$\mathcal{D_L}$ \%} & \multicolumn{5}{c|}{SH-A}                                                                                               & \multicolumn{5}{c}{UCF-QNRF}                                                                                           \\ \cline{2-11} 
		& \multicolumn{2}{c|}{No-GP ($\mathcal{D_L}$-only)} & \multicolumn{2}{c|}{GP ($\mathcal{D_L+D_U}$)} & \multirow{2}{*}{\begin{tabular}[c]{@{}c@{}}AG\\ \%\end{tabular}} & \multicolumn{2}{c|}{No-GP ($\mathcal{D_L}$-only)} & \multicolumn{2}{c|}{GP ($\mathcal{D_L+D_U}$)} & \multirow{2}{*}{\begin{tabular}[c]{@{}c@{}}AG\\ \%\end{tabular}} \\ \cline{2-5} \cline{7-10}
		& MAE          & MSE         & MAE        & MSE        &                                                                  & MAE          & MSE         & MAE        & MSE        &                                                                  \\ \hline
		5                                 & 118          & 211         & 102        & 172        & 16                                                               & 186          & 295         & 160        & 275        & 10                                                               \\ 
		25                                & 110          & 160         & 91         & 149        & 12                                                               & 178          & 252         & 147        & 226        & 14                                                               \\ 
		50                                & 102          & 149         & 89         & 148        & 6.1                                                              & 158          & 250         & 136        & 218        & 13                                                               \\ 
		75                                & 93           & 146         & 88         & 139        & 4.7                                                              & 139          & 240         & 129        & 210        & 9.8                                                              \\ \hline\hline
		100                               & 76           & 126         & -          & -          &                               -                                   & 114          & 195         & -          & -          &                                -                                  \\ \hline
	\end{tabular}
}
\end{table}

\begin{figure}[t!]	
	\centering	
	\includegraphics[width=0.24\linewidth]{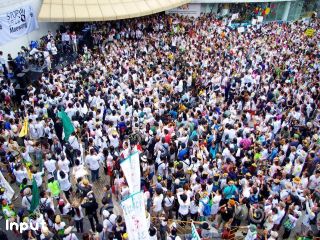}
	\includegraphics[width=0.24\linewidth]{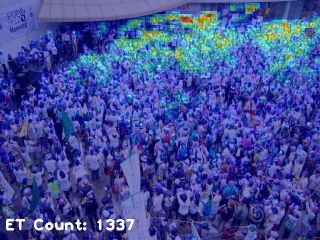}
	\includegraphics[width=0.24\linewidth]{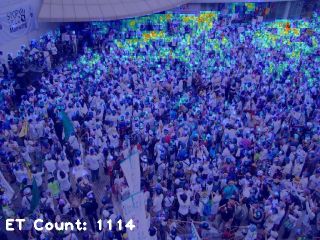}
	\includegraphics[width=0.24\linewidth]{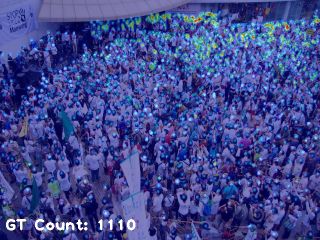}
	
	\includegraphics[width=0.24\linewidth]{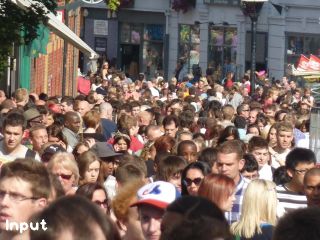}
	\includegraphics[width=0.24\linewidth]{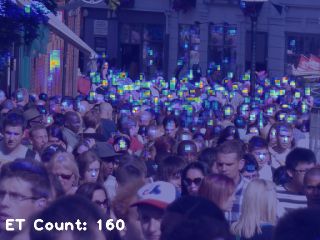}
	\includegraphics[width=0.24\linewidth]{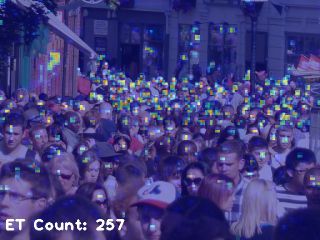}
	\includegraphics[width=0.24\linewidth]{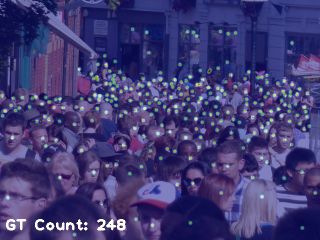}
	
	
	(a) \hskip 70pt (b) \hskip 70pt (c) \hskip 70pt (d) \hskip 70pt 
	\vskip -5pt\caption{Results of SSL experiments on the ShanghaiTech-A \cite{zhang2016single} dataset using the 5\% labeled data setting.  \textit{(a):} Input. \textit{(b)} No-GP  \textit{(c)} Proposed Method  \textit{(d)} Ground-truth.}
	\label{fig:ssl_shha}
\end{figure}

\begin{figure}[t!]	
	\centering	
	\includegraphics[width=0.24\linewidth]{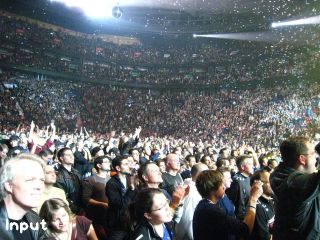}
	\includegraphics[width=0.24\linewidth]{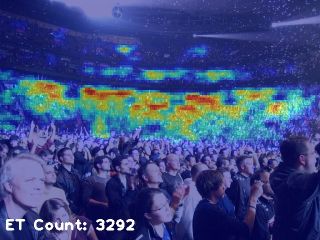}
	\includegraphics[width=0.24\linewidth]{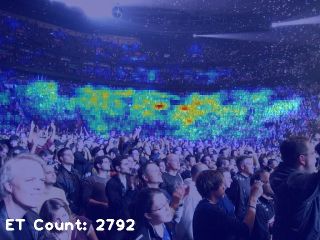}
	\includegraphics[width=0.24\linewidth]{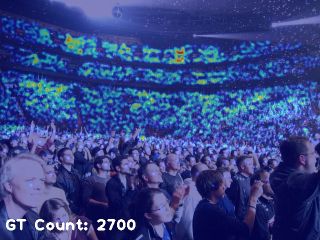}
	
	\includegraphics[width=0.24\linewidth]{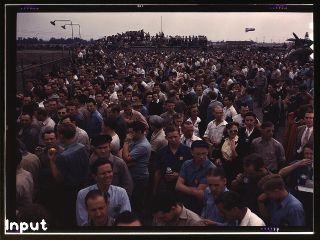}
	\includegraphics[width=0.24\linewidth]{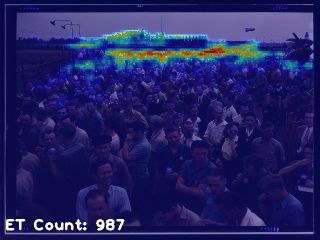}
	\includegraphics[width=0.24\linewidth]{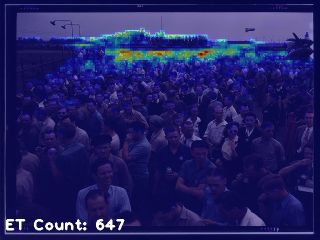}
	\includegraphics[width=0.24\linewidth]{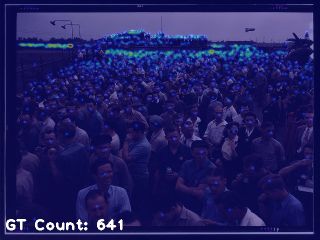}

	(a) \hskip 70pt (b) \hskip 70pt (c) \hskip 70pt (d) \hskip 70pt 
	\vskip -5pt\caption{Results of SSL experiments on the UCF-QNRF \cite{idrees2018composition} dataset using the 5\% labeled data setting.  \textit{(a):} Input. \textit{(b)} No-GP  \textit{(c)} Proposed Method  \textit{(d)} Ground-truth.}
	\label{fig:ssl_qnrf}
\end{figure}

\noindent\textbf{Ablation of labeled data percentage.} We conducted an ablation study where we varied the percentage of labeled data used during the training process. More specifically, we used $4$ different settings: 5\%, 25\%, 50\% and 75\%. The remaining data  were used as unlabeled samples. We used ResNet-50 as the network encoder for all the settings. This ablation study was conducted on 2 datasets: ShanghaiTech-A (SH-A) and UCF-QNRF. The results of this ablation study are shown in Table \ref{tab:ablation_lab}. It can be observed for both datasets that as the percentage of labeled data is reduced, the performance of the baseline network drops significantly. However, the proposed GP-based framework is able to leverage unlabeled data in all the cases to reduce this performance drop by a considerable margin. Fig. \ref{fig:ssl_shha} and \ref{fig:ssl_qnrf} show sample qualitative results on ShanghaiTech-A and UCF-QNRF datasets for the semi-supervised protocol with 5\% labeled data setting. It can be observed that the proposed method is able to predict the density maps more accurately as compared to the baseline method that does not consider unlabeled data. \\


\begin{table}[t!]
	\caption{Results of ablation study with different networks. The proposed method is able to exploit unlabeled data irrespective of different architectures. We used 5\% of the training data as labeled set, and the rest as unlabeled samples. AG: Average Gain \%$^{\ref{ftn:ag}}$. }
	\label{tab:ablation_arch}
	\vskip -9pt
	\resizebox{.81\linewidth}{!}{
		\begin{tabular}{l|c|cc|cc|c|cc|cc|c}
			\hline
			\multirow{3}{*}{Net}    & \multirow{3}{*}{$\mathcal{D_L}$\%} & \multicolumn{5}{c|}{SH-A}                                                                                                                                                                                                                                & \multicolumn{5}{c}{UCF-QNRF}                                                                                                                                                                                                                            \\ \cline{3-12} 
			&                                    & \multicolumn{2}{c|}{\begin{tabular}[c]{@{}c@{}}No-GP($\mathcal{D_L}$-only)\end{tabular}} & \multicolumn{2}{c|}{\begin{tabular}[c]{@{}c@{}}GP($\mathcal{D_L+D_U}$)\end{tabular}} & \multirow{2}{*}{\begin{tabular}[c]{@{}c@{}}AG\\ \%\end{tabular}} & \multicolumn{2}{c|}{\begin{tabular}[c]{@{}c@{}}No-GP  ($\mathcal{D_L}$-only)\end{tabular}} & \multicolumn{2}{c|}{\begin{tabular}[c]{@{}c@{}}GP ($\mathcal{D_L+D_U}$)\end{tabular}} & \multirow{2}{*}{\begin{tabular}[c]{@{}c@{}}AG\\ \%\end{tabular}} \\ \cline{3-6} \cline{8-11}
			&                                    & MAE                                          & MSE                                          & MAE                                        & MSE                                        &                                                                  & MAE                                          & MSE                                          & MAE                                        & MSE                                        &                                                                  \\ \hline
			\multirow{2}{*}{ResNet-50}  & 100                                & 76                                           & 126                                          & -                                          & -                                          & -                                                                & 114                                          & 195                                          & -                                          &                                            & -                                                                \\ 
			& 5                                  & 118                                          & 211                                          & 102                                        & 172                                        & 16                                                               & 186                                          & 295                                          & 160                                        & 275                                        & 10                                                               \\ \hline\hline
			\multirow{2}{*}{ResNet-101} & 100                                & 76                                           & 117                                          & -                                          & -                                          & -                                                                & 116                                          & 197                                          & -                                          & -                                          & -                                                                \\
			& 5                                  & 131                                          & 200                                          & 110                                        & 162                                        & 18                                                               & 196                                          & 324                                          & 174                                        & 288                                        & 11                                                               \\ \hline\hline
			\multirow{2}{*}{VGG16}  & 100                                & 74                                           & 118                                          & -                                          & -                                          & -                                                                & 120                                          & 197                                          & -                                          &                                            & -                                                                \\ 
			& 5                                  & 121                                          & 205                                          & 112                                        & 163                                        & 14                                                               & 188                                          & 316                                          & 175                                        & 291                                        & 7.4                                                              \\ \hline
		\end{tabular}
	}
\end{table}

\noindent\textbf{Architecture ablation.} We conducted an ablation study where we evaluated the proposed method using different architectures.  More specifically, we used different networks like ResNet-50, ResNet-101 and VGG16 as encoder network. The ablation was performed on 2 datasets: ShanghaiTech-A (SH-A) and UCF-QNRF.  For all the experiments, we used 5\% of the training dataset as labeled dataset, and the rest were used as unlabeled samples. The results of this experiment are shown in Table \ref{tab:ablation_arch}. Based on these results, we make the following observations: (i) Since networks like VGG16 and ResNet-101 have higher number of parameters, they tend to overfit more in the reduced-data setting as compared to ResNet-50.  (ii)  The proposed GP-based method obtains consistent gains by leveraging unlabeled dataset across different architectures.\\

\begin{figure}[h!]
	\begin{center}
		\includegraphics[width=0.7\linewidth]{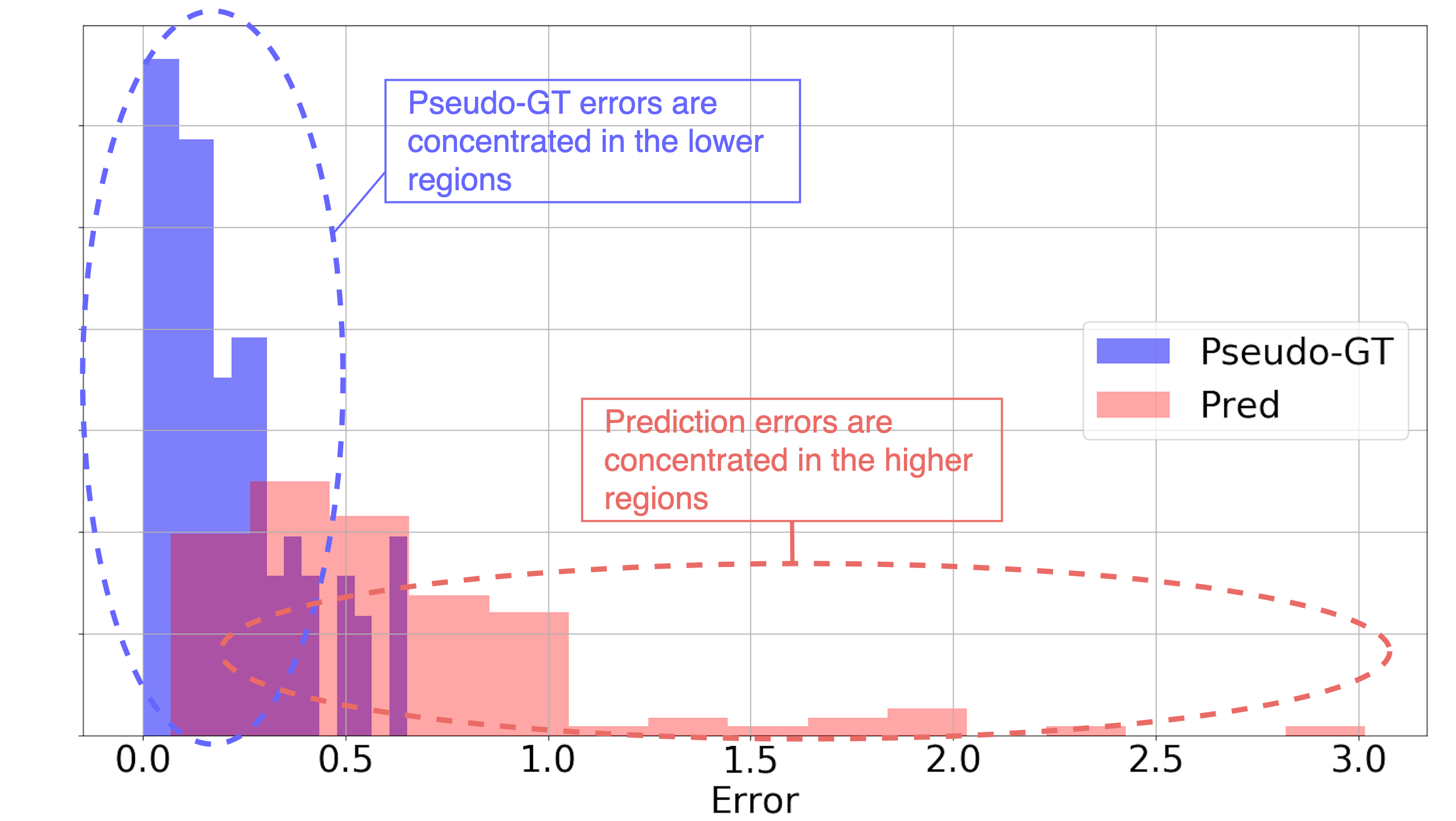} 
	\end{center}
	\vskip -15pt
	\caption{Histogram  for pseudo-GT errors ($err_{pseudo}^u$) and prediction errors ($err_{pred}^u$) on unlabeled data during training. Note that  pseudo-GT errors are concentrated on the lower end, implying that they are more closer to the ground truth as compared to the predictions.   Hence, pseudo-GTs provide meaningful supervision.}
	\label{fig:pseudo_gt_analysis}
\end{figure}

\noindent\textbf{Pseudo-GT Analysis.} In order to gain a deeper understanding about the effectiveness of the proposed approach, we plot the histogram of normalized  errors with respect to the predictions  $y_{pred}^u$ of the network and the pseudo-GT $y_{pseudo}^u$ for the unlabeled data during the training process. Specifically, we plot histograms of $err_{pred}^u$ and $err_{pseudo}^u$, where $err_{pred}^u = \frac{|y_{pred}^u - y_{gt}^u|}{y_{gt}^u}$ and $err_{pseudo}^u = \frac{|y_{pseudo}^u - y_{gt}^u|}{y_{gt}^u}$. Here, $y_{gt}^u$ is the actual GT corresponding to the unlabeled data sample. The plot is shown in Fig. \ref{fig:pseudo_gt_analysis}. It can be observed that the pseudo-GT errors are concentrated in the lower end of the error region as compared to the prediction errors. This implies that the pseudo-GTs are more closer to the GTs than the predictions. Hence, the pseudo-GTs obtained using the proposed method are able to provide good quality supervision on the unlabeled data.

\subsection{Synthetic-to-Real transfer setting}
Recently, Wang \etal \cite{wang2019learning} proposed a synthetic crowd counting dataset (GCC) that consists of 15,212 images with a total of  7,625,843 annotations. The primary purpose of this dataset is to reduce the annotation efforts by training the networks on the synthetic dataset, thereby eliminating the need for labeling. However, due to a   gap between the synthetic and real-world data distributions, the networks trained on synthetic dataset perform poorly on real-world images. In order to overcome this issue, the authors in \cite{wang2019learning}  proposed a Cycle-GAN based domain adaptive approach that additionally enforces SSIM consistency. More specifically, they first learn to translate from synthetic crowd images to real-world images using SSIM-based Cycle-GAN. This transfers the style in the synthetic image to more real-world style. The translated synthetic images are then used to train a  counting network (SFCN) that is based on ResNet-101 architecture. 

While this approach improves the error over the baseline methods, its performance is essentially limited in the case of large distribution gap between real and synthetic images. Moreover, the authors in \cite{wang2019learning} perform a manual selection of synthetic samples for training the network. This selections ensures that only  samples that are closer to the real-world images in terms of the count are used for training. Such a selection is not feasible in the case of unsupervised domain adaptation where we have no access to  labels in the target dataset.

\begin{table}[t!]
	\caption{Comparison of results in synthetic-to-real transfer settings. We train the network on synthetic crowd counting dataset (GCC), and leverage the training set of real-world datasets without any labels. We used the same network as described in \cite{wang2019learning}.} 
	\label{tab:da}
	\vskip -10pt
	\resizebox{.81\linewidth}{!}{
	\begin{tabular}{l|cc|cc|cc|cc|c}
		\hline
		\multirow{2}{*}{Method}              & \multicolumn{2}{c|}{SH-A} & \multicolumn{2}{c|}{SH-B} & \multicolumn{2}{c|}{UCF-QNRF} & \multicolumn{2}{c|}{UCF-CC-50} & WExpo \\ \cline{2-10} 
		& MAE         & MSE         & MAE         & MSE         & MAE           & MSE           & MAE            & MSE           & MAE   \\ \hline
		No Adapt                             & 160         & 217         & 22.8        & 30.6        & 276           & 459           & 487            & 689           & 42.8  \\ 
		Cycle GAN \cite{zhu2017unpaired}                  & 143         & 204         & 24.4        & 39.7        & 257           & 401           & 405            & 548           & 32.4  \\ 
		SE Cycle GAN \cite{wang2019learning} & 123         & 193         & 19.9        & 28.3        & 230           & 384           & 373            & 529           & 26.3  \\
		Proposed Method                      & \textbf{121}         & \textbf{181}         & \textbf{12.8}        & \textbf{19.2}        & \textbf{210}           & \textbf{351}           & \textbf{355}            & \textbf{505}           & \textbf{20.4}  \\ \hline
	\end{tabular}
}
\end{table}

\begin{figure}[t!]	
	\centering	
	\includegraphics[width=0.24\linewidth]{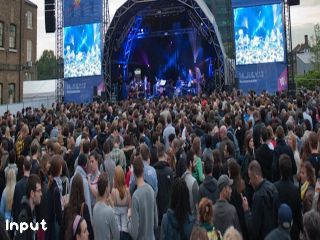}
	\includegraphics[width=0.24\linewidth]{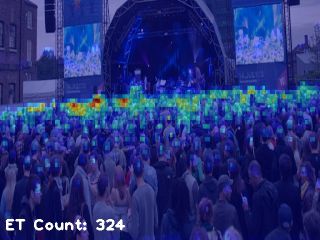}
	\includegraphics[width=0.24\linewidth]{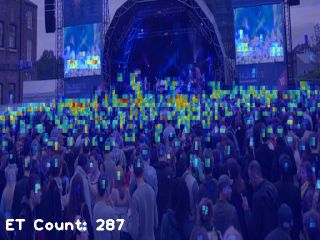}
	\includegraphics[width=0.24\linewidth]{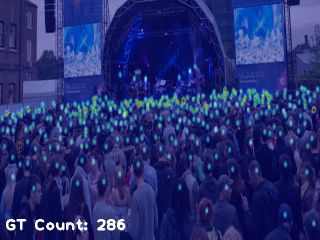}
	
	\includegraphics[width=0.24\linewidth]{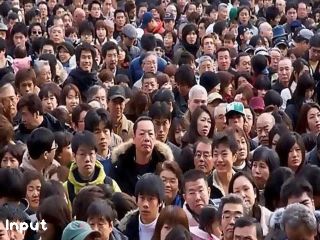}
	\includegraphics[width=0.24\linewidth]{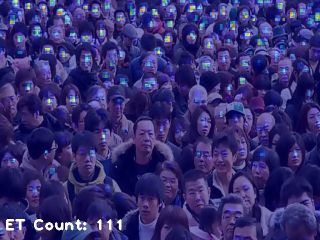}
	\includegraphics[width=0.24\linewidth]{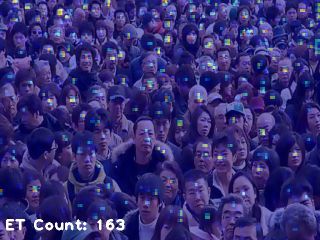}
	\includegraphics[width=0.24\linewidth]{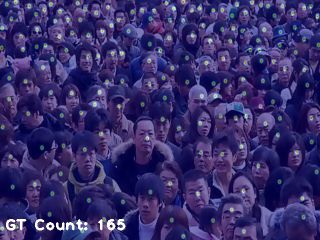}	\\
	\vskip -3pt
	(a) \hskip 70pt (b) \hskip 70pt (c) \hskip 70pt (d) \hskip 70pt 
	\vskip -5pt\caption{Results of Synthetic-to-Real transfer experiments on ShanghaiTech-A dataset.  \textit{(a):} Input. \textit{(b)} No Adapt \textit{(c)} Proposed Method  \textit{(d)} Ground-truth.}
	\label{fig:da_shha}	
\end{figure}

\begin{figure}[t!]	
	\centering	
	\includegraphics[width=0.24\linewidth]{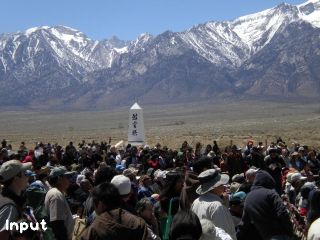}
	\includegraphics[width=0.24\linewidth]{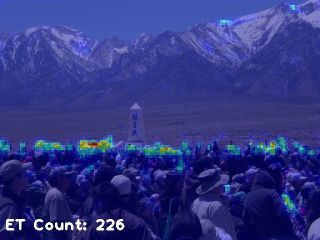}
	\includegraphics[width=0.24\linewidth]{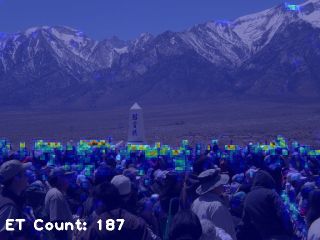}
	\includegraphics[width=0.24\linewidth]{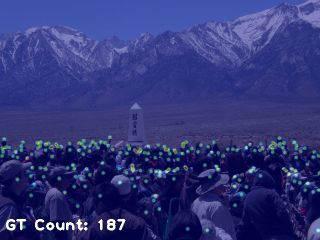}
	
	\includegraphics[width=0.24\linewidth]{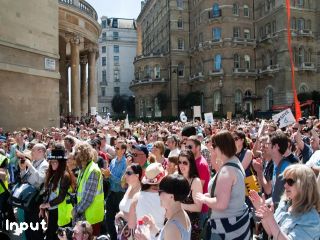}
	\includegraphics[width=0.24\linewidth]{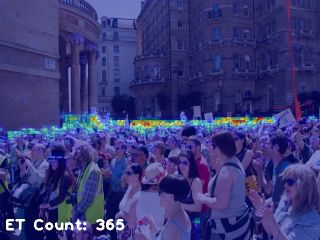}
	\includegraphics[width=0.24\linewidth]{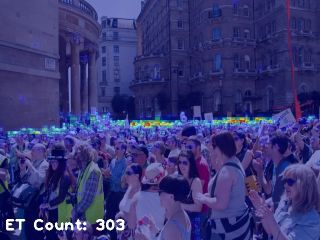}
	\includegraphics[width=0.24\linewidth]{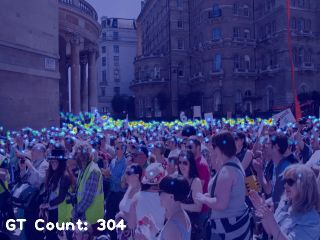}	\\
	\vskip -3pt
	(a) \hskip 70pt (b) \hskip 70pt (c) \hskip 70pt (d) \hskip 70pt 
	\vskip -5pt\caption{Results of Synthetic-to-Real transfer experiments on the UCF-QNRF \cite{idrees2018composition} dataset.  \textit{(a):} Input. \textit{(b)} No Adapt. \textit{(c)} Proposed Method.  \textit{(d)} Ground-truth.}
	\label{fig:da_qnrf}	
\end{figure}

The proposed GP-based framework overcomes these drawbacks easily and can be extended to the synthetic-to-real transfer  setting as well. We consider the synthetic data as labeled training set and real-world training set as unlabeled dataset, and train the network to leverage the unlabeled dataset. The results of this experiment are reported in Table \ref{tab:da}. We used the same network (SFCN) and training process as described in \cite{wang2019learning}.  As it can be observed, the proposed method achieves considerable improvements compared to the recent approach. Since we estimate the pseudo-GT for unlabeled real-world images and use it as supervision directly, the distribution gap that the network needs to handle is much lesser. This results in better performance compared to the domain adaptive approach \cite{wang2019learning}. Unlike \cite{wang2019learning}, we train the network on the unlabeled data and hence,  we do not need to perform any synthetic sample selection. Fig. \ref{fig:da_shha} and \ref{fig:da_qnrf} show sample qualitative results on the ShanghaiTech-A and UCF-QNRF datasets for the synthetic-to-real transfer  protocol. The proposed method is able to predict the density maps more accurately as compared to the baseline.

\section{Conclusions}
In this work, we focused on learning to count in the crowd from limited labeled data.  Specifically, we proposed a GP-based iterative learning framework that involves estimation of pseudo-GT for unlabeled data using Gaussian Processes, which is then used as supervision for training the network. Through various experiments, we show that the proposed method can be effectively used in a variety of scenarios that involve unlabeled data like learning with less data or synthetic to real-world transfer. In addition, we conducted detailed ablation studies to demonstrate that the proposed method generalizes well to different network architectures and is able to achieve consistent gains for different amounts of labeled data.

\section*{Acknowledgement}

This work was supported by the NSF grant 1910141.

\section*{Supplementary Material}

Due to limited space in the main paper, we present additional details about the proposed method and experiments in the supplementary.

\subsection*{Encoder and Decoder Architecture}
Here, we provide details of the encoder and decoder architecture for all the   experiments. \\

\noindent\textbf{Encoder:} In the main paper, we conducted experiments with 4 different networks for the encoder: For semi-supervised experiments, we used Res50, Res101 and VGG16. For learning from synthetic data we used Res101-SFCN \cite{wang2019learning} .Following are the details:\\

\noindent(i) Res50: First 3 layers of Res50 are used as the encoder.

\noindent(ii) Res101:  First 3 layers of Res101 are used as the encoder.

\noindent(iii) VGG16: First 10 layers of VGG16 are used as the encoder.

\noindent(iv) Res101-SFCN: We use the network exactly as described in \cite{wang2019learning}. In this network, the layers until final dilated conv layer are considered as a part of  the encoder.\\

For  all the above networks, the features of the final encoder layer are forwarded through a $1\times1$ conv layer to reduce the dimensionality to 64 channels. The output of this $1\times1$ conv is the feature embedding in the latent space which is used in GP modeling. Since the train crop size is $256\times256$, the intermediate feature maps  in the latent space is of dimension $64\times32\times32$. \\

\noindent\textbf{Decoder:} We use the same decoder in all the semi-supervised learning experiments. The decoder consists of 2 conv-relu layers. The first one is a $3\times3$ conv layer, that takes in 64 channels and outputs 64 channels. The final layer is a a $1\times1$ layer that takes in 64 channels and outputs 1 channel which is the density map. The final conv layer is  followed by an bilinear-upsampling layer that upsamples the output density to the resolution of the input image. 

In case of learning from the synthetic data, since we use the same network as in \cite{wang2019learning}, all the layers after the dilated conv layers are used as decoder.

\subsection*{Dataset Details}

In this section, we provide details of the different datasets used for evaluating the proposed method in the main paper. \\

\noindent\textbf{ShanghaiTech \cite{zhang2016single}}:This dataset  contains 1198 annotated images with a total of 330,165 people. This dataset consists of two parts: Part A with 482 images and Part B with 716 images. Both parts are further divided into training and test datasets with training set of Part A containing 300 images and that of Part B containing 400 images. Rest of the images are used as test set.\\

\noindent\textbf{UCF-QNRF \cite{idrees2018composition}}: UCF-QNRF  is a large crowd counting dataset with 1535 high-resolution images and 1.25 million head annotations. There are 1201 training images and 334 test images. It contains extremely congested scenes where the maximum count of an image can reach 12865.\\

\noindent\textbf{WorldExpo \cite{zhang2015cross}}: The WorldExpo’10 dataset was introduced by Zhang \etal. \cite{zhang2015cross} and it contains 3,980 annotated frames from 1,132 video sequences captured by 108 surveillance cameras. The frames are divided into training and test sets. The training set contains 3,380 frames and the test set contains 600 frames from five different scenes with 120 frames per scene. They also provided Region of Interest (ROI) map for each of the five scenes.\\

\noindent\textbf{UCSD \cite{chan2008privacy}}: The UCSD dataset crowd counting dataset consists of 2000 frames from a single scene. These scenes contain relatively sparse crowds with the number of people ranging from 11 to 46 per frame. A region of interest (ROI) is pro- vided for the scene in the dataset.  Of the 2000 frames, frames 601 through 1400 are used for training while the remaining frames are held out for testing. \\

\noindent\textbf{GCC \cite{wang2019learning}}:GTA V Crowd Counting Dataset (GCC) is a large-scale synthetic dataset based on an electronic game, which consists of 15,212 crowd images. GCC provides three evaluation strategies (random splitting, cross-camera,and cross-location evaluation).\\

\begin{table}[htp!]
	\caption{Effect of $\lambda_{un}$ on ShanghaiTech Part-A val set.}
	\label{tab:lambda}
	\begin{tabular}{l|cc}
		\hline
		$\lambda_{un}$ & MAE & MSE \\ \hline
		0.0            & 102 & 175 \\ 
		0.2            & 100 & 162 \\
		0.4            & 89  & 149 \\ 
		0.6            & 85  & 140 \\ 
		0.8            & 88  & 147 \\ 
		1.0            & 92  & 156 \\ \hline
	\end{tabular}
\end{table}

\subsection*{Hyper-parameter $\lambda_{un}$}

In this section, we study the effect of $\lambda_{un}$ on the overall performance.   $\lambda_{un}$  weighs the unsupervised loss function in the Eq. 12 of main paper. For this study, we use the ShanghaiTech A dataset, due to its wide variety of scenes and diversity in the count. We conducted this experiment for the 5\% data setting where 5\% of the data was used as labeled data and rest was used as unlabeled data. We used Res50 encoder. Note that we perform the evaluation on the held-out validation set (and not on the test set). The results for different values of $\lambda_{un}$ are shown in Table \ref{tab:lambda}.

We observed that the performance peaks when the value of $\lambda_{un}$ is $0.6$.  $\lambda_{un}=0$ corresponds to only labeled data. This is the baseline performance. As we increase $\lambda_{un}$, we observe that the error improves. However, for $\lambda_{un}  > 0.6$, we see a small drop. This is because the network would not have learned to optimal level at the initial stages of training. Due to this the pseud-GT will be erroneous, and hence, using high weight for unsupervised at initial stages prohibits the network from reaching optimal performance. 

Based on this experiment, we use $\lambda_{un} = 0.6$ for all the experiments.

\begin{table}[t!]
	\caption{Semi-supervised experiments with recent crowd counting methods. We used 5\% of the training data as labeled set, and the rest as unlabeled samples. AG: Average Gain \%$^{\ref{ftn:ag}}$. }
	\label{tab:ssl_recent}
	\vskip -9pt
	\resizebox{.9\linewidth}{!}{
		\begin{tabular}{l|c|cc|cc|c|cc|cc|c}
			\hline
			\multirow{3}{*}{Net}    & \multirow{3}{*}{$\mathcal{D_L}$\%} & \multicolumn{5}{c|}{SH-A}                                                                                                                                                                                                                                & \multicolumn{5}{c}{UCF-QNRF}                                                                                                                                                                                                                            \\ \cline{3-12} 
			&                                    & \multicolumn{2}{c|}{\begin{tabular}[c]{@{}c@{}}No-GP($\mathcal{D_L}$-only)\end{tabular}} & \multicolumn{2}{c|}{\begin{tabular}[c]{@{}c@{}}GP($\mathcal{D_L+D_U}$)\end{tabular}} & \multirow{2}{*}{\begin{tabular}[c]{@{}c@{}}AG\\ \%\end{tabular}} & \multicolumn{2}{c|}{\begin{tabular}[c]{@{}c@{}}No-GP  ($\mathcal{D_L}$-only)\end{tabular}} & \multicolumn{2}{c|}{\begin{tabular}[c]{@{}c@{}}GP ($\mathcal{D_L+D_U}$)\end{tabular}} & \multirow{2}{*}{\begin{tabular}[c]{@{}c@{}}AG\\ \%\end{tabular}} \\ \cline{3-6} \cline{8-11}
			&                                    & MAE                                          & MSE                                          & MAE                                        & MSE                                        &                                                                  & MAE                                          & MSE                                          & MAE                                        & MSE                                        &                                                                  \\ \hline
			\multirow{2}{*}{Res101-SFCN} & 100                                & 74                                           & 114                                          & -                                          & -                                          & -                                                                & 113                                          & 196                                          & -                                          & -                                          & -                                                                \\
			& 5                                  & 128                                          & 199                                          & 109                                        & 160                                        & 17                                                               & 193                                          & 323                                          & 172                                       & 282                                        & 12                                                               \\ \hline\hline
			\multirow{2}{*}{CSRNet}  & 100                                & 71                                           & 112                                          & -                                          & -                                          & -                                                                & 123                                          & 195                                         & -                                          &                                            & -                                                                \\ 
			& 5                                  & 120                                          & 200                                          & 111                                        & 159                                        & 14                                                               & 187                                          & 310                                          & 171                                        & 293                                       & 7.0                                                             \\ \hline
		\end{tabular}
	}
\end{table}

\begin{table}[t!]
	
	\centering
	\caption{Results in SSL settings.  Reducing labeled data to 5\% results in performance drop by a big margin as compared to 100\% data. Res50 was used as the encoder network for all the methods. RL: Ranking-Loss. GP: Gaussian-Process. AG: Average Gain \%$^{\ref{ftn:ag}}$.}
	\label{tab:tab1_std}
	\resizebox{1\linewidth}{!}{
		\begin{tabular}{l|c|c|c|c|c|c|c|c|c|c|c|c|c|c|c|c} 
			\hline
			\multirow{2}{*}{Method} & \multirow{2}{*}{$\mathcal{D_L}$ } & \multirow{2}{*}{$\mathcal{D_U}$ } & \multicolumn{3}{c|}{SH-A}            & \multicolumn{3}{c|}{SH-B}              & \multicolumn{3}{c|}{UCF-QNRF}        & \multicolumn{2}{c|}{WExpo} & \multicolumn{3}{c}{UCSD}             \\ 
			\cline{4-17}
			&                                   &                                   & MAE           & MSE           & AG & MAE            & MSE            & AG& MAE           & MSE           & AG & MAE            & AG     & MAE           & MSE           & AG \\ 
			\hline
			Ours          & 5\%                               &           95\%                         & {102 $\pm$ 0.8}  & {172 $\pm$ 2.1}  & {16}    & {15.7 $\pm$ 0.9}  & {27.9 ($\pm$ 1.1)}  & {22}     & {160 $\pm$ 2.4}  & {275 $\pm$ 3.1}  &  {10}     & {12.8 $\pm$ 0.5}  & {10}         & {2.0 $\pm$ 0.05}  & {2.4 $\pm$ 0.09}  &  {12}     \\
			\hline
		\end{tabular}
	}
\end{table}

\begin{table}[t!]
	\caption{Results for synthetic-to-real transfer settings. We train the network on synthetic crowd counting dataset (GCC), and leverage the training set of real-world datasets without any labels. We used the same network and training/evaluation protocol as in \cite{wang2019learning}.} 
	\label{tab:tab4_std}
	\vskip -10pt
	\resizebox{1\linewidth}{!}{
		\begin{tabular}{l|c|c|c|c|c|c|c|c|c}
			\hline
			\multirow{2}{*}{Method}              & \multicolumn{2}{c|}{SH-A} & \multicolumn{2}{c|}{SH-B} & \multicolumn{2}{c|}{UCF-QNRF} & \multicolumn{2}{c|}{UCF-CC-50} & WExpo \\ \cline{2-10} 
			& MAE         & MSE         & MAE         & MSE         & MAE           & MSE           & MAE            & MSE           & MAE   \\ \hline
			Ours                     & {121 $\pm$ 0.6}         & {181 $\pm$ 1.6}         & {12.8$\pm$ 0.3}        & {19.2$\pm$ 0.9}        & {210$\pm$ 2.7}           & {351$\pm$ 4.1}           & {355 $\pm$ 4.4}            & {505$\pm$ 5.9}           & {20.4 $\pm$ 0.9}  \\ \hline
		\end{tabular}
	}
\end{table}

\subsection*{Additional Architecture Ablation}

In this section, we conducted additional architecture ablation experiments using two recent crowd counting techniques: CSRNet \cite{li2018csrnet} and Res101-SFCN \cite{wang2019learning}. WE use the 5\% data-setting, where we use 5\% of the data as labeled and rest as unlabeled. We evaluated both these methods on ShanghaiTech-A (SH-A) and UCF-QNRF datasets.For CSRNet, we use the layers upto the last dilated conv as the encoder. For the decoder, we use 2 conv layers as described earlier. 

The results of this experiment are shown in Table \ref{tab:ssl_recent}. In addition to MAE/MSE, we rerport Average Gain (AG)\footnote{\label{ftn:ag}$AG=\frac{G_{mae}+G_{mse}}{2}$, $G_{mae} =  \frac{mae_{(\mathcal{D_U+D_L})} - mae_{(\mathcal{D_L})}}{mae_{(\mathcal{D_L})}}$, $G_{mse} =  \frac{mse_{(\mathcal{D_U+D_L})} - mse_{(\mathcal{D_L})}}{mse_{(\mathcal{D_L})}}$}.
We observed consistent gains in both the cases when we used the proposed GP-based method to leverage unlabeled data.

\subsection*{Multiple Trials}

In this section, we report the standard-deviations for the experiments with our proposed method corresponding to Table 1 and Table 4 in the main paper. See Table \ref{tab:tab1_std} and Table \ref{tab:tab4_std}. Note that the standard deviations are computed using 5 trials.

%
%
\bibliographystyle{splncs04}
\bibliography{egbib}
\end{document}